\documentclass{article}

\usepackage[final,nonatbib]{nips_2016}

\usepackage[utf8]{inputenc} 
\usepackage[T1]{fontenc}    
\usepackage{url}            
\usepackage{booktabs}       
\usepackage{amsfonts}       
\usepackage{nicefrac}       
\usepackage{microtype}      
\usepackage{graphicx}
\usepackage{listings}
\usepackage{courier}
\usepackage{subcaption}

\newcommand{\pchoose}{\texttt{choose}}
\newcommand{\porganism}{\texttt{organism}}
\newcommand{\peat}{\texttt{eat}}

\lstdefinelanguage{scala}{
  morekeywords={abstract,case,catch,class,def,%
    do,else,extends,false,final,finally,%
    for,if,implicit,import,match,mixin,%
    new,null,object,override,package,%
    private,protected,requires,return,sealed,%
    super,this,throw,trait,true,try,%
    type,val,var,while,with,yield},
  otherkeywords={=>,<-,<\%,<:,>:,\#,@},
  sensitive=true,
  morecomment=[l]{//},
  morecomment=[n]{/*}{*/},
  morestring=[b]'',
  morestring=[b]',
  morestring=[b]"''
}
\title{Probabilistic Neural Programs}

\author{
  Kenton W. Murray \thanks{Work done while on Internship at Allen Institute for Artificial Intelligence} \\
  Department of Computer Science and Engineering\\
  University of Notre Dame\\
  South Bend, IN 46617 \\
  \texttt{kmurray4@nd.edu} \\
  \And
  Jayant Krishnamurthy \\
  Allen Institute for Artificial Intelligence \\
  Seattle, WA 98166 \\
   \texttt{jayantk@allenai.org} \\
}

\begin{document}

\maketitle

\begin{abstract}
We present probabilistic neural programs, a framework for program
induction that permits flexible specification of both a computational
model and inference algorithm while simultaneously enabling the use of
deep neural networks. 
Probabilistic neural programs combine a computation graph for
specifying a neural network with an operator for weighted
nondeterministic choice. Thus, a program describes both a collection
of decisions as well as the neural network architecture used to make
each one. We evaluate our approach on a challenging diagram question
answering task where probabilistic neural programs correctly execute
nearly twice as many programs as a baseline model.
\end{abstract}


%



\section{Introduction}

In recent years, deep learning has produced tremendous accuracy
improvements on a variety of tasks in computer vision and natural
language processing. A natural next step for deep learning is to
consider \emph{program induction}, the problem of learning computer
programs from (noisy) input/output examples. Compared to more
traditional problems, such as object recognition that require making
only a single decision, program induction is difficult because it
requires making a sequence of decisions and possibly learning control
flow concepts such as loops and if statements.

Prior work on program induction has described two general classes of
approaches. First, in the noise-free setting, program synthesis
approaches pose program induction as completing a program ``sketch,''
which is a program containing nondeterministic choices (``holes'') to
be filled by the learning algorithm
\cite{solar-lezama2006}. Probabilistic programming languages
generalize this approach to the noisy setting by permitting the sketch
to specify a \emph{distribution} over these choices as a function of
prior parameters and further to condition this distribution on data,
thereby training a Bayesian generative model to execute the sketch
correctly \cite{GMR08}. Second, neural abstract machines define
continuous analogues of Turing machines or other general-purpose
computational models by ``lifting'' their discrete state and
computation rules into a continuous representation
\cite{NeelakantanLS15,ReedF15,graves2014neural,riedel2016programming}. Both
of these approaches have demonstrated success at inducing simple
programs from synthetic data but have yet to be applied to practical
problems.

We observe that there are (at least) three dimensions along which we
can characterize program induction approaches:

\begin{enumerate}
\item Computational Model -- what abstract model of computation does the
  model learn to control? (e.g., a Turing machine)
\item Learning Mechanism -- what kinds of machine learning models are
  supported? (e.g., neural networks, Bayesian generative models)
\item Inference -- how does the approach reason about the many
  possible executions of the machine?
\end{enumerate}

Neural abstract machines conflate some of these dimensions: they
naturally support deep learning, but commit to a particular
computational model and approximate inference algorithm. These choices
are suboptimal as (1) the bias/variance trade-off suggests that
training a more expressive computational model will require more data
than a less expressive one suited to the task at hand, and (2) recent
work has suggested that discrete inference algorithms may outperform
continuous approximations \cite{gaunt2016}. In contrast, probabilistic
programming supports the specification of different (possibly
task-specific) computational models and inference algorithms,
including discrete search and continuous approximations. However,
these languages are restricted to generative models and cannot
leverage the power of deep neural networks.

We present probabilistic neural programs, a framework for program
induction that permits flexible specification of the computational
model and inference algorithm while simultaneously enabling the use of
deep neural networks. Our approach builds on computation graph
frameworks \cite{abadi2016,bergstra2010theano} for specifying neural
networks by adding an operator for weighted nondeterministic choice
that is used to specify the computational model. Thus, a program
sketch describes \emph{both} the decisions to be made and the
architecture of the neural network used to score these
decisions. Importantly, the \emph{computation graph interacts with
  nondeterminism}: the scores produced by the neural network determine
the weights of nondeterministic choices, while the choices determine
the network's architecture. As with probabilistic programs, various
inference algorithms can be applied to a sketch. Furthermore, a
sketch's neural network parameters can be estimated using stochastic
gradient descent from either input/output examples or full execution
traces.

We evaluate our approach on a challenging diagram question answering
task, which recent work has demonstrated can be formulated as learning
to execute a certain class of probabilistic programs. On this task, we
find that the enhanced modeling power of neural networks improves
accuracy.

%
%
%
%
%
%
%
%
%
%
%
%

\section{Probabilistic Neural Programs}

Probabilistic neural programs build on computation graph frameworks
for specifying neural networks by adding an operator for
nondeterministic choice. We have developed a Scala library for
probabilistic neural programming that we use to illustrate the key
concepts.

\begin{figure}
\begin{subfigure}[b]{0.45\textwidth}
\begin{lstlisting}[language=scala]
def mlp(v: Tensor):
  Pp[CgNode] =
  for {
    w1 <- param("w1")
    b1 <- param("b1")
    h1 = ((w1 * v) + b1).tanh
    w2 <- param("w2")
    b2 <- param("b2")
    out = (w2 * h1) + b2
  } yield {
    out
  }
\end{lstlisting}
\end{subfigure}
\begin{subfigure}[b]{0.55\textwidth}
\begin{lstlisting}[language=scala]
val dist: Pp[Int] = for {
  s <- mlp(new Tensor(...))
  v <- choose(Array(0, 1), s)
  y <- choose(Array(2, 3), s)
} yield {
  v + y
}
// tensor parameters initialized to 0
val params: NnParams
println(dist.beamSearch(10, params))
// output: 2 (0.25), 3 (0.25),
//         3 (0.25), 4 (0.25)
\end{lstlisting}
\end{subfigure}
%
\vspace{-.3in}
\caption{Probabilistic neural programs defining a multilayer
  perceptron with a computation graph (left) and applying it to create a probability
  distribution over program executions (right).}
\label{fig:pnp}
\vspace*{-\baselineskip}
\end{figure}

Figure \ref{fig:pnp} (left) defines a multilayer perceptron as a
probabilistic neural program. This definition resembles those of other
computation graph frameworks. Network parameters and intermediate
values are represented as computation graph nodes with tensor
values. They can be manipulated with standard operations such as
matrix-vector multiplication and hyperbolic tangent. Evaluating this
function with a tensor yields a program sketch object that can be
evaluated with a set of network parameters to produce the network's
output.

Figure \ref{fig:pnp} (right) shows how to use the \pchoose{} function
to create a nondeterministic choice. This function
nondeterministically selects a value from a list of options. The score
of each option is given by the value of a computation graph node that
has the same number of elements as the list. Evaluating this function
with a tensor yields a program sketch object that represents a
function from neural network parameters to a \emph{probability
  distribution} over values. The log probability of a value is
proportional to the sum of the scores of the choices made in the
execution that produced it. Performing (approximate) inference over
this object -- in this case, using beam search -- produces an explicit
representation of the distribution. Multiple nondeterministic choices
can be combined to produce more complex sketches; this capability can
be used to define complex computational models, including
general-purpose models such as Turing machines. The library also has
functions for conditioning on observations.

Although various inference algorithms may be applied to a program
sketch, in this work we use a simple beam search over executions. This
approach accords with the recent trend in structured prediction to
combine greedy inference or beam search with powerful non-factoring
models \cite{parsey,nivre2007,dyer2016}. The beam search maintains a
queue of partial program executions, each of which is associated with
a score. Each step of the search continues each execution until it
encounters a call to \pchoose{}, which adds zero or more executions to
the queue for the next search step. The lowest scoring executions are
discarded to maintain a fixed beam width. As an execution proceeds, it
may generate new computation graph nodes; the search maintains a
single computation graph shared by all executions to which these nodes
are added. The search simultaneously performs the forward pass over
these nodes as necessary to compute scores for future choices.

The neural network parameters are trained to maximize the
loglikelihood of correct program executions using stochastic gradient
descent. Each training example consists of a pair of program sketches,
representing an unconditional and conditional distribution. The
gradient computation is similar to that of a loglinear model with
neural network factors. It first performs inference on both the
conditional and unconditional distributions to estimate the expected
counts associated with each nondeterministic choice. These counts are
then backpropagated through the computation graph to update the
network parameters.

\section{Diagram Question Answering with Probabilistic Neural Programs}

\begin{figure}[t]
  \centering
  \begin{subfigure}[b]{.50\textwidth}
  \fbox{\includegraphics[scale=0.50]{2111.png}\rule[0cm]{0cm}{0cm} \rule[0cm]{0cm}{0cm}}
\end{subfigure}
\begin{subfigure}[b]{.45\textwidth}
What happens to the snake population if the field mouse population decreases?

$\lambda f.\texttt{cause}(\texttt{decrease}(\texttt{mice}),f(\texttt{snakes}))$
\\

Does the deer eat the grass?

$\texttt{eats}(\texttt{deer}, \texttt{grass})$
\\

How many organisms eat the grass?

$\texttt{count}(\lambda x.\texttt{eats}(x, \texttt{grass}))$
\\

Are rabbits a tertiary consumer?

$\texttt{tertiary-consumer}(\texttt{rabbits})$

\end{subfigure}
\vspace{-.05in}
\caption{A food web diagram with annotations generated from a computer vision system (left) along with related questions and their associated program sketches (right).}
\label{foodweb}
\vspace*{-\baselineskip}
\end{figure}




We consider the problem of learning to execute program sketches in a
food web computational model using visual information from a
diagram. This problem is motivated by recent work \cite{EMNLP16},
which has demonstrated that diagram question answering can be
formulated as translating natural language questions to program
sketches in this model, then learning to execute these sketches.
Figure \ref{foodweb} shows some example questions from this work,
along with the accompanying diagram that must be interpreted to
determine the answers. The diagram (left) is a food web, which depicts
a collection of organisms in an ecosystem with arrows to indicate what
each organism eats. The right side of the figure shows questions
pertaining to the diagram and their associated program sketches.

%
%
%


The possible executions of each program sketch are determined by a
domain-specific computational model that is designed to reason about
food webs. The nondeterministic choices in this model correspond to
information that must be extracted from the diagram. Specifically,
there are two functions that call \pchoose{} to nondeterministically
return a boolean value. The first function, \porganism{}($x$), should
return true if the text label $x$ is an organism (as opposed to e.g.,
the image title). The second function, \peat{}$(x,y)$, should return
true if organism $x$ eats organism $y$. These functions do influence
program control flow. The food web model also
includes various other functions, e.g., for reasoning about population
changes, that call \porganism{} and \peat{} to extract information
from the diagram. \cite{EMNLP16} has a more thorough description of
the theory; our goal is to learn to make the choices in this theory.



We consider three models for learning to make the choices for both
\porganism{} and \peat{}: a non-neural (\textsc{Loglinear}) model, as
well as two probabilistic neural models (\textsc{2-Layer PNP} and
\textsc{Maxpool PNP}). All three learn models for both \porganism{} and \peat{}
using outputs from a computer vision system trained to detect
organism, text, and arrow relations between them.
\cite{EMNLP16} defines a set of hand-engineered features heuristically
created from the outputs of this vision system. \textsc{Loglinear}
and \textsc{2-Layer PNP} use only these features, and the difference
is simply in the greater expressivity of a two-layer neural network.
However, one of the major strengths of neural models is
their ability to learn latent feature representations automatically, 
and our third model also uses the direct outputs of the vision system
not made into features. The architecture of \textsc{Maxpool PNP} reflects
this and contains additional input layers that maxpool over detected
relationships between objects and confidence scores.
The expectation is that our neural network
modeling of nondeterminism will learn better latent representations
than the manually defined features.

\section{Experiments}



We evaluate probabilistic neural programs on the \textsc{foodwebs}
dataset introduced by \cite{EMNLP16}.  This data set contains a
training set of \textasciitilde 2,900 programs and a test set of
\textasciitilde 1,000 programs. These programs are human annotated
gold standard interpretations for the questions in the data set, which
corresponds to assuming that the translation from questions to
programs is perfect. We train our probabilistic neural programs using
correct execution traces of each program, which are also provided in
the data set.

%

We evaluate our models using two metrics.  First, execution accuracy
measures the fraction of programs in the test set that are executed
completely correctly by the model. This metric is challenging because
correctly executing a program requires correctly making a number of
\pchoose{} decisions.  Our 1,000 test programs had over 35,000
decisions, implying that to completely execute a program correctly
means getting on average 35 \pchoose{} decisions correct without
making any mistakes. Second, \pchoose{} accuracy measures the accuracy
of each decision independently, assuming all previous decisions were
made correctly.


\begin{table}
\centering
\begin{tabular}[h]{ l c c }
\hline
 & Execution & \pchoose{} \\
Method & Accuracy & Accuracy \\
\hline
\textsc{Loglinear} & 8.6\% & 78.2\% \\
\textsc{2-Layer PNP} & 12.5\% & 78.7\% \\
\textsc{Maxpool PNP} & \textbf{14.9\%} & \textbf{82.5\%} \\
\hline
\end{tabular}
\vspace{2 pt}
\caption{Results from appling Probabilistic Neural Programs to the
  execution of sketches from the \textsc{foodwebs}
  dataset. ``\pchoose{} Accuracy'' indicates average accuracy at each
  decision, whereas ``Execution Accuracy'' indicates if the entire
  program was executed correctly.}
\label{sequences}
\vspace*{-.3in}
\end{table}

Table \ref{sequences} compares the accuracies of our 
three models on the \textsc{foodwebs} dataset.
The improvement in accuracy between the baseline (\textsc{Loglinear})
and the probabilistic neural program (\textsc{2-Layer PNP}) is due to the 
neural network's enhanced modeling power.
Though the \pchoose{} accuracy does not improve by a large margin, 
the improvements translate into large gains in entire program correctness.
Finally, as expected, the inclusion of lower level features
(\textsc{Maxpool PNP}) not possible in \textsc{Loglinear} significantly
improved performance. Note that this task
requires performing computer vision, and thus it is not expected that
any model achieve 100\% accuracy.

\section{Conclusion}

We have presented probabilistic neural programs, a framework for
program induction that permits flexible specification of computational
models and inference algorithms while simultaneously enabling the use
of deep learning. A program sketch describes a collection of
nondeterministic decisions to be made during execution, along with the
neural architecture to be used for scoring these decisions. The
network parameters of a sketch can be trained from data using
stochastic gradient descent. We demonstrate that probabilistic neural
programs improve accuracy on a diagram question answering task which
can be formulated as learning to execute program sketches in a
domain-specific computational model.

\section*{Acknowledgements}

The authors would like to thank the reviewers for their comments as
well as helpful discussions with Arturo Argueta and Oyvind Tafjord.

\medskip

\small

%
%

\bibliographystyle{plain}
\bibliography{neural_probabilistic_programs}

\end{document}